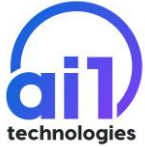 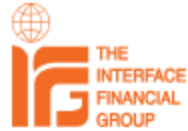

# Predicting Invoice Dilution in Supply Chain Finance with Leakage-Free Two-Stage Models: XGBoost, KAN (Kolmogorov – Arnold Networks), and Ensembles

*Pavel Koptev[1], Vishnu Kumar[2], Dr. Konstantin Malkov[1], George Shapiro[2], Yury Vikhanov[1]*

## Abstract

Invoice or payment dilution—the gap between the approved invoice amount and the actual collection—is a significant source of non-credit risk and margin loss in supply chain finance. Traditionally, this risk is managed through the buyer's irrevocable payment undertaking (IPU), which commits to full payment without deductions. However, IPUs can hinder supply chain finance adoption, particularly among sub-invested grade buyers. A newer, data-driven methods use real-time dynamic credit limits, projecting dilution for each buyer-supplier pair in real-time. This paper introduces a machine learning framework and evaluates how that can supplement a deterministic algorithm to predict invoice dilution using extensive production dataset across nine key transaction fields.

The ScoreAI pipeline enriches these fields with leakage-free historical time-series statistics (computed only from invoices strictly prior to the current invoice date) and macroeconomic indicators, and trains a leakage-free, two-stage architecture for realistic deployment. Stage 1 uses an XGBoost binary classifier to detect whether an invoice is likely to dilute; in **Machine Learning Analysis Report** rolling-window evaluation it achieves ROC-AUC between 0.9167 and 0.9222 (0.9222 on the 2021–2025 time-window holdout window) and average precision between 0.8361 and 0.8498. Stage 2 estimates dilution magnitude using several regressor families (XGBoost, RandomForest, MLP, FasterKAN) and ensemble variants; across seven rolling time-window holdout windows, the weighted ensemble achieves mean RMSE 1215.8 ± 99.7 with WMAPE 16.82% ± 0.48 (MAPE values are ~17–19% by model). We also report an ablation study to quantify the incremental impact of macroeconomic features.

## 1 Introduction

In any receivables-based finance, invoice or payment dilution refers to the reduction in realized payment relative to the original invoice amount (Bakker et al., 2004; Klapper, 2006). Unlike in the invoice or receivable financing, due to the fact that invoices to be funded are not necessary approved yet for payment by account debtors/buyers but rather verified, where major reasons for dilutions are related to issue with invoice underlaying deliverables, in SCF dilution arises from buyer's debit entries, volume discounts, counterclaims and other post-approved payment adjustments. For supply-chain finance providers, unanticipated post-approval dilution directly reduces yield on funded assets and can turn nominally profitable transactions into losses.

---

[1] AI1 Technologies,
[2] The Interface Financial Group

The data employed in this paper were obtained from The Interface Financial Group (IFG), a financial services provider that specializes in supply chain financing through the provision of early payment mechanisms for invoices issued by suppliers to buyers operating across multiple jurisdictions and industrial sectors. Its production data now comprises millions of invoice-level transactions, each represented in a compact schema of nine features. Historically, in receivable finance dilution risk has been managed through conservative advance rates and static reserves, but in SCF the risk has been managed by requiring buyer's guarantee or IPU (Camerinelli, 2009). However, reliance on IPU requirement, is shifting the risk solely to the unsecured credit risk of the buyer and limits the SCF service to investment grade or near-investment grade buyers. IFG solved this issue by creating deterministic algorithms based on data-driven real time dynamic credit limits for buyer-supplier pairs, which among other parameters includes historical dilution multipliers. Although these heuristics effectively mitigate tail risk, their performance can be enhanced through the integration of predictive AI models. Also, the predictive AI model previously developed was unable to significantly contribute to the improvement.

The authors aimed to enhance the accuracy and effectiveness of a predictive model designed for invoice-level dilution. The goal is to assess whether integrating the model as an extra contributor to the real-time dynamic credit limit underwriting approach can provide an additional forward-looking dilution estimate for every proposed transaction. The improved method designed to deliver better result than previous ML model. The previous reviewed by authors model was based on initial analysis which was conducted using the Orange data-mining toolkit (developed by the University of Ljubljana, Slovenia) (Demšar et al., 2013), evaluating algorithms such as k-nearest neighbors, support vector machines, random forest, neural networks, and AdaBoost for regression modeling. Further examination with K-Nearest Neighbors (k-NN) and neural networks in Python followed.

## 2 Dataset and Preprocessing

The underlying production dataset contains millions of historical transactions, each linking a supplier to a specific buyer and capturing the approved invoice amount, total payments received, and relevant dates. In its core schema, each invoice record includes nine transaction fields (buyer_id, supplier_id, invoice_number, currency, approved invoice amount, total_payment_amount, issue_date, payment_date, dilution_amount). In the experiments summarized here, the full dataset contains 4,836,699 records with 137 input columns after feature engineering:

To avoid label leakage, the ScoreAI pipeline computes historical behavior features using only prior invoices by date. These leakage-free statistics are generated for each Buyer and each Buyer–Supplier pair across multiple lookback horizons (e.g., 180/360/720 days and all-time), including dilution frequency, typical dilution size (amounts and percentages), and recency signals (days since last invoice / last dilution). The model also incorporates macroeconomic indicators (unemployment, PCE, GDP, industrial production index, and retail sales) to provide broader context.

The dataset was partitioned into seven distinct time periods utilizing a four-year sliding window approach. Each segment was then subdivided, with 70% designated for training, 15% reserved for validation during hyperparameter tuning, and the remaining 15% allocated for testing to assess performance metrics. Subsequently, a sequence of seven experiments was conducted to evaluate the final model.

## 3 Models

We adopt a leakage-free, multi-level approach that reflects how dilution risk is used operationally: first detect dilution risk, then estimate its magnitude only when risk is present. For magnitude estimation, we evaluate several regressor families (XGBoost, RandomForest, MLP, and FasterKAN) and ensemble combinations to improve robustness across time intervals.

Stage 1 — XGBoost Classifier. A binary classifier predicts whether an invoice will experience any dilution. This stage supports early warning, triage, and threshold-based policy decisions (e.g., additional review, adjusted reserves, or tighter limits).

Stage 2 — Regression Models. A regression model predicts the expected dilution amount for invoices flagged as likely diluted by Stage 1. In **Machine Learning Analysis Report** we train multiple candidate regressors (XGBoost, RandomForest, MLP, and FasterKAN) with hyperparameter optimization. This separation improves interpretability and avoids forcing a single model to learn both event detection and impact estimation.

Stage 3 — Ensemble (AVG/WGT). To leverage complementary model strengths, we also construct simple-average and weighted-average ensembles over the Stage 2 regressors and report their rolling-window performance alongside individual models.

## 4 Experimental Results

To measure robustness over time, we divided the full dataset into training and test subsets to collect statistics on model quality across different calendar date ranges. We then recalculated (retrained) the Stage 1 classifier and Stage 2 regressors for each range, compiled summary diagnostics, and generated additional plots and diagrams. To evaluate the model's reliability, several experiments using a randomization approach were conducted, demonstrating the model's strong performance. Finally, we ran a controlled ablation in which macroeconomic parameters were removed to quantify their impact.

| Component | Metric | Value | Notes |
|---|---|---|---|
| Stage 1: XGBoost classifier | ROC-AUC (time-window holdout; range, mean) | 0.9167–0.9222 (0.9205) | Stable across rolling windows; avg precision 0.8361–0.8498. |

| Stage 2: MLP regressor | MAPE (time-window holdout; mean ± std) | 17.84% ± 0.33 | Best single-model average; see Table 2 for full comparison. |
| Stage 3: Ensemble regressor (WGT) | RMSE (time-window holdout; mean ± std) | 1215.8 ± 99.7 | Best overall; weighted ensemble average WMAPE = 16.82% ± 0.48. |

Stage 1 (dilution-event classification) is stable across time windows: ROC-AUC ranges from 0.9167 to 0.9222 (mean 0.9205) and average precision ranges from 0.8361 to 0.8498 (mean 0.8436). Per-class recalls from the normalized confusion matrices vary by operating threshold and window (Non-diluted: 79.01–88.68%; Diluted: 74.20–94.30%). Table 1 summarizes the rolling-window classifier diagnostics from **Machine Learning Analysis Report**

*Table 1 – Stage 1 classifier diagnostics across rolling calendar windows (time-window holdout).*

| Calendar window | ROC-AUC | Avg. precision | Recall (Non-diluted) | Recall (Diluted) |
|---|---|---|---|---|
| 2015-2019 | 0.9218 | 0.8457 | 79.01% | 94.30% |
| 2016-2020 | 0.9222 | 0.8498 | 88.05% | 77.45% |
| 2017-2021 | 0.9202 | 0.8428 | 82.72% | 88.61% |
| 2018-2022 | 0.9167 | 0.8493 | 84.43% | 83.14% |
| 2019-2023 | 0.9200 | 0.8446 | 88.64% | 74.20% |
| 2020-2024 | 0.9204 | 0.8361 | 80.95% | 90.31% |
| 2021-2025 | 0.9222 | 0.8369 | 88.68% | 75.67% |

For Stage 2 magnitude estimation, **Machine Learning Analysis Report** compares multiple regressor families (XGBoost, RandomForest, MLP, FasterKAN) and ensemble variants across seven rolling time-window holdout windows. Table 2 summarizes the mean ± standard deviation of regression metrics across windows. Ensembles improve over any single model; the weighted ensemble delivers the best overall RMSE (1215.8 ± 99.7) and WMAPE (16.82% ± 0.48) on average.

*Table 2 – Stage 2 regressor metrics by model across rolling calendar windows (time-window holdout; mean ± std).*

| Model | R² | MAE | RMSE | MAPE (%) | WMAPE (%) |
|---|---|---|---|---|---|
| XGBoost | 0.8433 ±0.0212 | 297.0914 ±9.9873 | 1372.5669 ±85.9816 | 18.0758 ±0.7138 | 24.5048 ±1.0166 |
| RandomForest | 0.8585 ±0.0233 | 289.1397 ±8.3774 | 1302.6359 ±101.0308 | 17.7901 ±0.6583 | 23.8512 ±0.9621 |
| FasterKAN | 0.8481 ±0.0323 | 306.6415 ±14.8216 | 1347.7169 ±143.2914 | 19.1279 ±0.4736 | 25.2876 ±1.2554 |
| MLP | 0.8632 ±0.0234 | 285.0667 ±0.0234 | 1281.3949 ±110.5764 | 17.8387 ±0.3303 | 23.5136 ±0.9712 |
| **Ensemble (AVG)** | 0.8712 ±0.0200 | 283.7568 ±9.2897 | 1243.5975 ±90.8746 | **17.2399 ±0.4146** | 16.9059 ±0.5411 |
| **Ensemble (WGT)** | **0.8768 ±0.0209** | **282.3884 ±8.7000** | **1215.8378 ±99.6911** | 17.5501 ±0.3922 | **16.8238 ±0.4842** |

Impact of macroeconomic parameters: to isolate their contribution, we trained an otherwise identical baseline XGBoost model with and without macroeconomic indicators and evaluated on time-window holdout. As shown in Table 3, the macroeconomic features provide a modest improvement across metrics (within ~1%), suggesting that richer buyer- and supplier-specific external signals may further improve performance.

*Table 3 – Impact of macroeconomic parameters on model performance (time-window holdout).*

| Model | R² | MAE | RMSE | MAPE (%) | WMAPE (%) |
|---|---|---|---|---|---|
| XGBoost (with macroeconomic parameters) | 0.7983 | 406.8951 | 1580.9263 | 25.7995 | 33.1412 |
| XGBoost (without macroeconomic parameters) | 0.7969 | 408.6923 | 1586.2864 | 25.8211 | 33.2876 |

## 4.1 Visual Diagnostics

Figures below reproduce key diagnostics from **Machine Learning Analysis Report**. To keep the narrative concrete, we show plots for the most recent 2021–2025 evaluation window (time-window holdout), which is representative of the overall stability observed across all rolling windows. Stage 2 visual diagnostics below correspond to the best-performing Stage 2 model for the 2021–2025 window (typically the weighted ensemble), while Table 2 summarizes the broader Stage 2 model and ensemble comparison across windows.

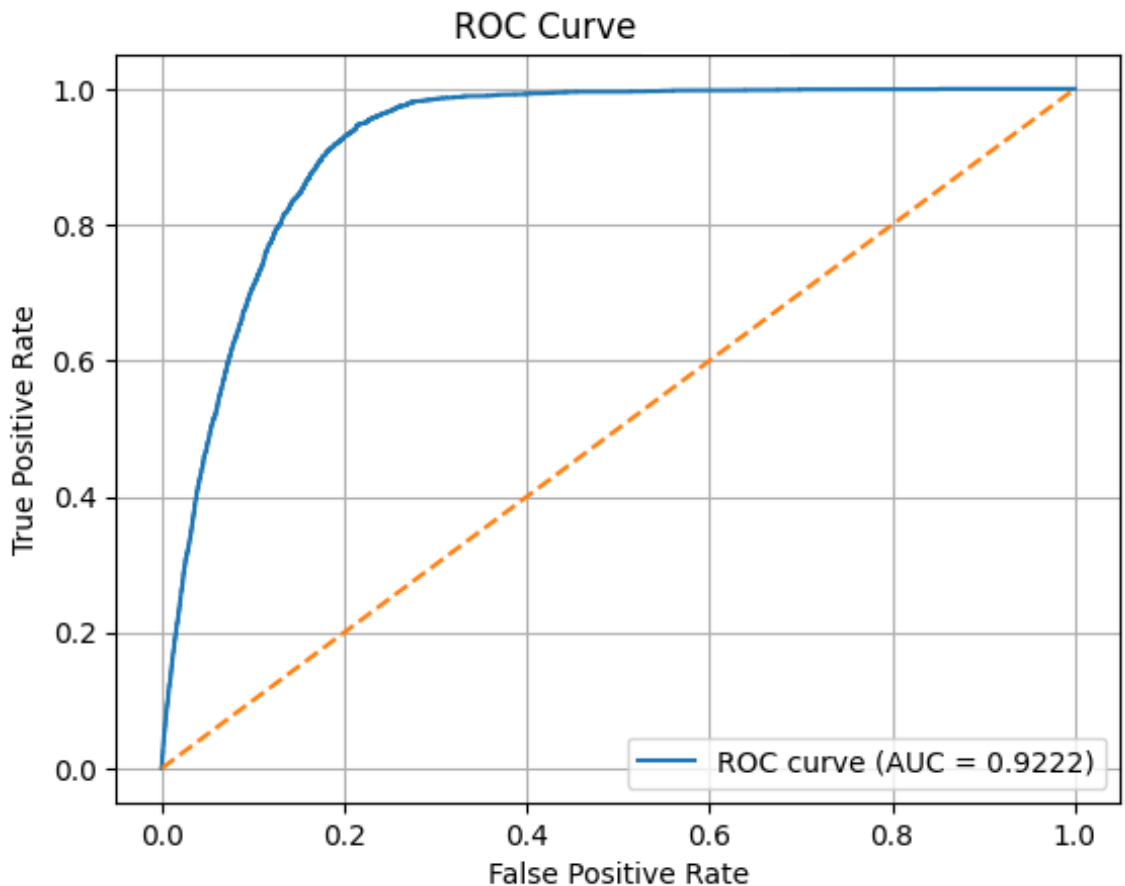

*Figure 1 – ROC curve for the Stage 1 classifier (time-window holdout example: 2021–2025; AUC = 0.9222).*

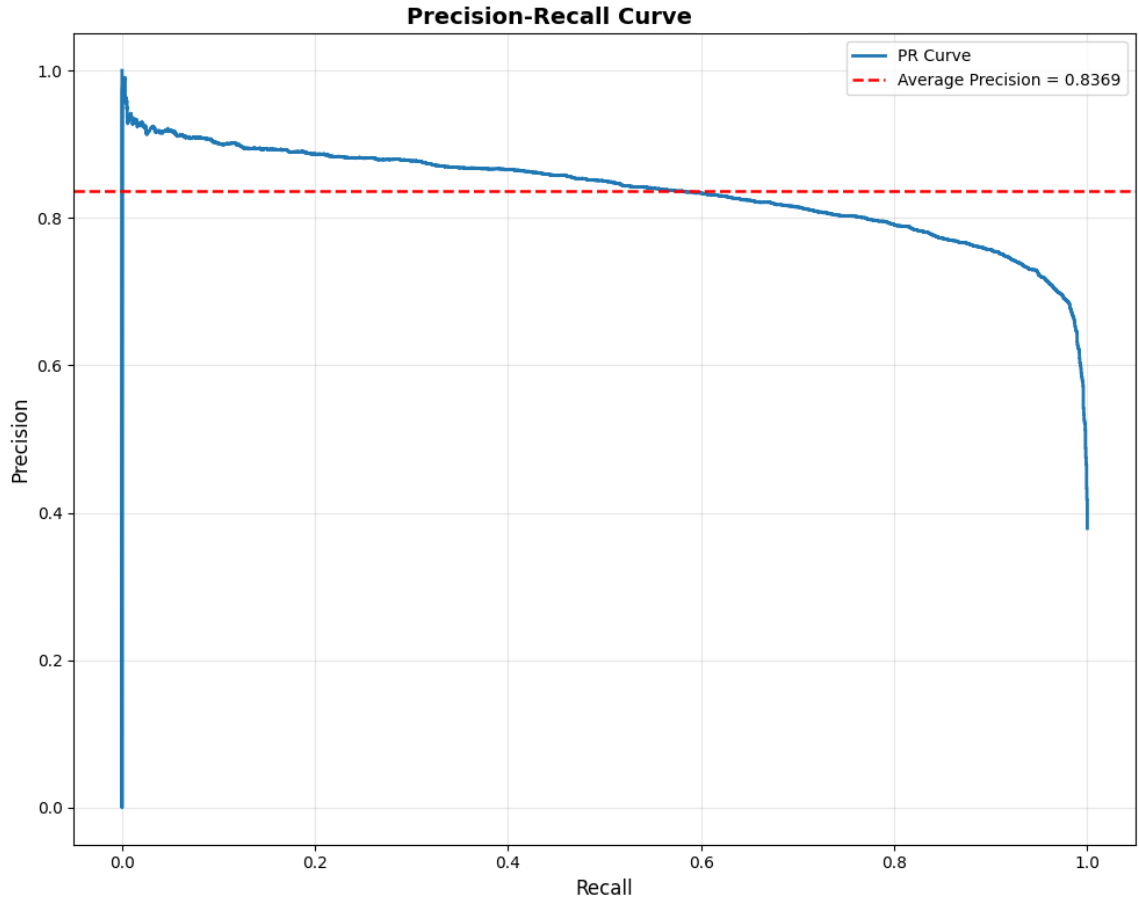

*Figure 2 – Precision–Recall curve for the Stage 1 classifier (time-window holdout example: 2021–2025; average precision = 0.8369).*

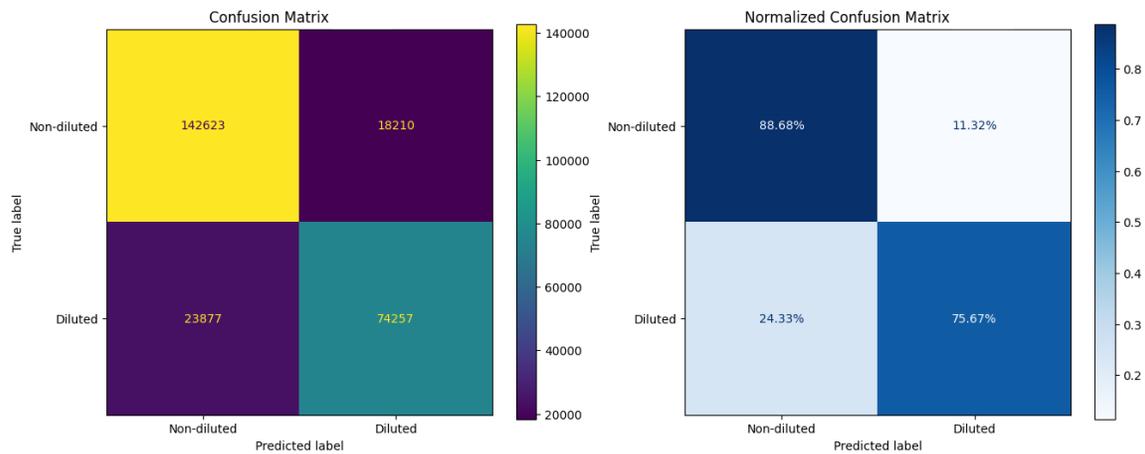

*Figure 3 – Confusion matrix for the Stage 1 classifier (time-window holdout example: 2021–2025; normalized).*

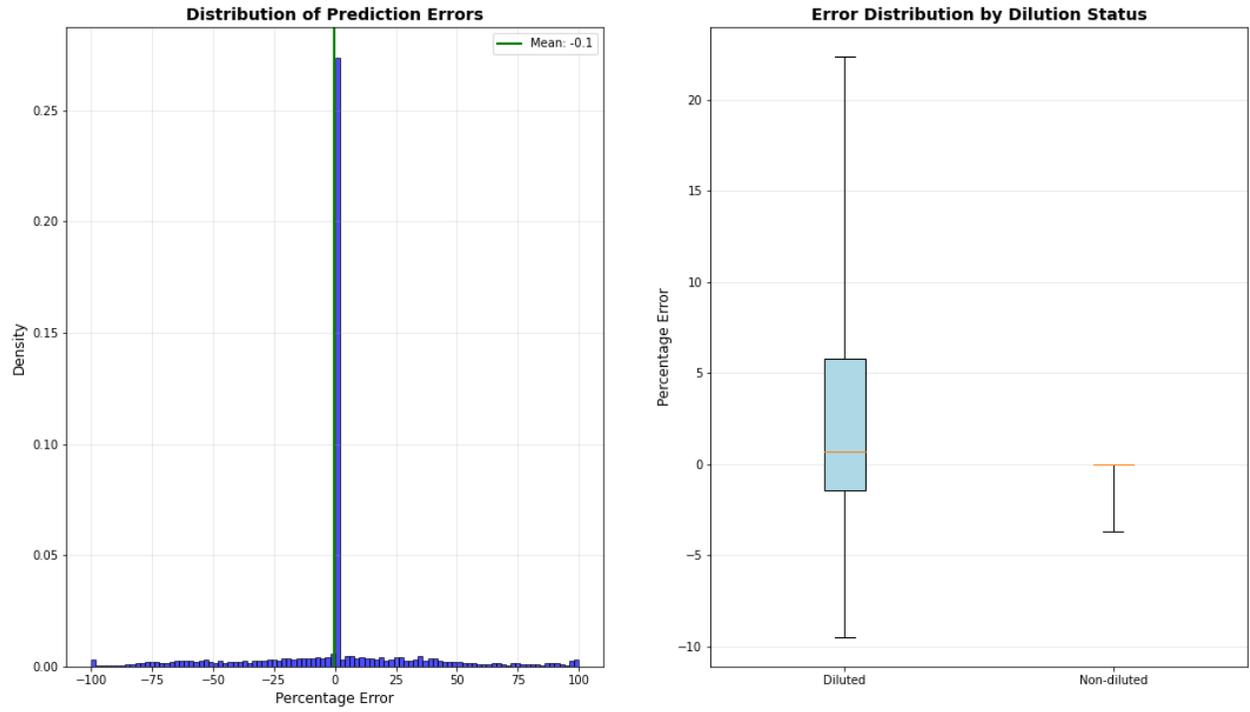

*Figure 4 – Distribution of prediction errors and error distribution by dilution status (Test Set, 2021–2025).*

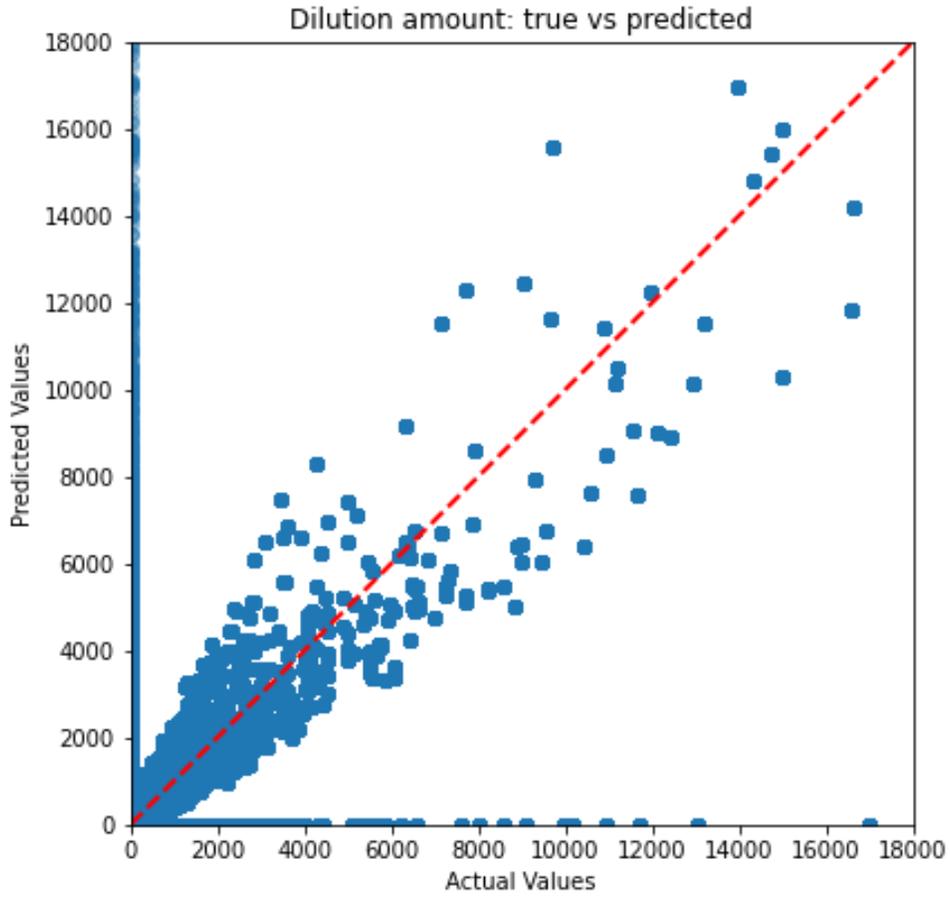

*Figure 5 – Two-level model: actual vs. predicted dilution amount on Test Set (2021–2025).*

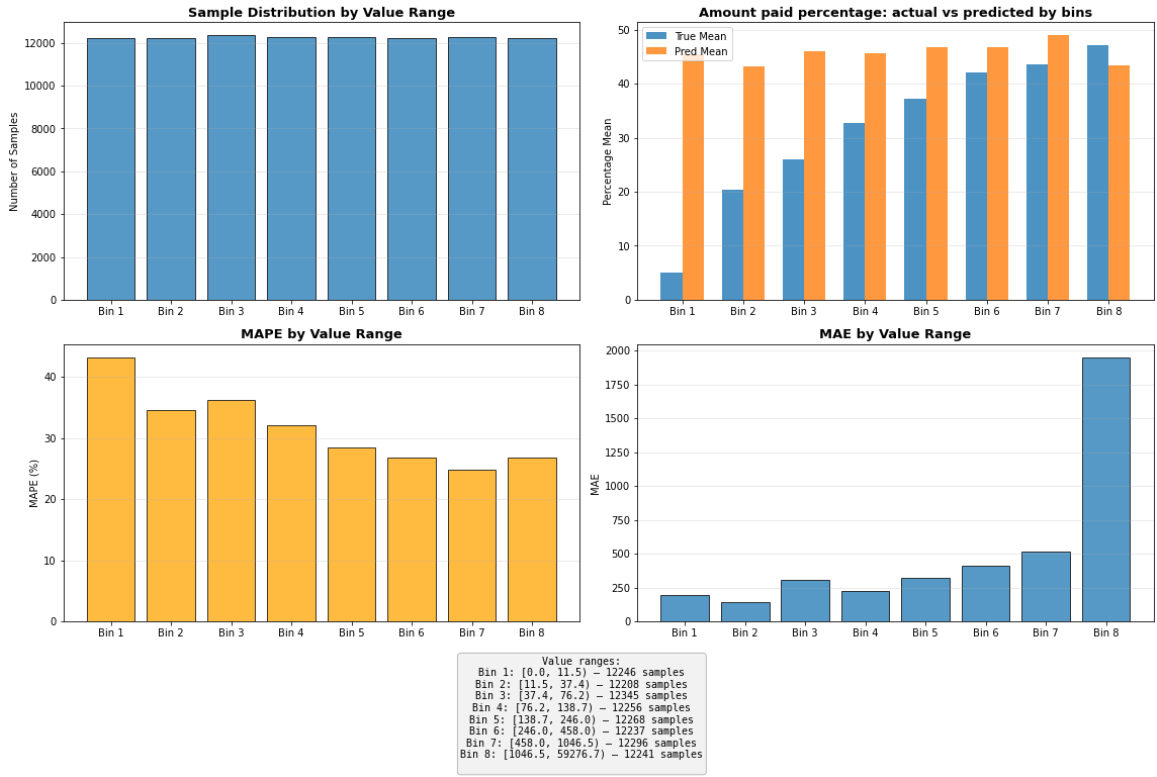

*Figure 6 – Model performance analysis by target value range (Test Set, 2021–2025): sample distribution, mean calibration, MAPE, and MAE by bin.*

## 5 Discussion

The **Machine Learning Analysis Report** results support a practical operating model: first predict whether dilution risk is present (Stage 1), then estimate its magnitude only when risk is present (Stage 2). Across rolling windows, ensembles provide a consistent improvement over single model regressors; the weighted ensemble achieves the best overall average RMSE (1215.8 ± 99.7) and WMAPE (16.82% ± 0.48) (Table 2). A controlled ablation indicates that macroeconomic indicators offer modest additional lift (<1% across metrics; Table 3).

Stage 1 maintains strong separation performance (ROC-AUC = 0.9222 on the 2021–2025 time-window holdout window; 0.9167–0.9222 across all rolling windows). The confusion matrix in Figure 3 illustrates one threshold-dependent operating point; in practice, the operating threshold should be tuned to business costs (e.g., cost of review vs. cost of surprise dilution).

Stage 2 amount prediction exhibits the expected heavy-tail behavior: errors are modest for the bulk of invoices but increase for extreme outcomes. The binned analysis in Figure 6 shows that MAPE is highest for the smallest dilution-value ranges and decreases for larger ranges, while MAE increases with magnitude. The two-level actual-versus-predicted plot (Figure 5) shows broad alignment with the diagonal but increasing dispersion in the tails—suggesting that high-value tail events may benefit from separate handling (e.g., specialized models, quantile estimates, or policy-based caps).

## 6 Conclusion and Soon to Come

We have presented an updated machine learning framework for predicting invoice and payment dilution in supply chain finance. The **Machine Learning Analyss Report** approach uses leakage-free, time-ordered historical features and a multi-level design: an XGBoost classifier to detect dilution risk and a suite of candidate regressors (XGBoost, RandomForest, MLP, FasterKAN) with optional ensemble combinations to estimate expected dilution magnitude when risk is present.

On held-out time-window test data, the classifier achieves ROC-AUC = 0.9222 on the most recent (2021–2025) evaluation window, with ROC-AUC ranging 0.9167–0.9222 across rolling windows. This provides a robust basis for operational early warning and prioritization. For magnitude estimation, rolling-window comparisons show that ensembles outperform individual regressors; the weighted ensemble achieves mean RMSE 1215.8 ± 99.7 with WMAPE 16.82% ± 0.48 across windows (Table 2). An ablation study indicates that macroeconomic indicators provide a modest additional lift (Table 3).

Next steps to improve amount accuracy and reduce uncertainty include: (i) continuing to refine class-imbalance handling without discarding majority-class records (e.g., cost-sensitive learning or other balancing approaches), (ii) extending ensemble methods to the Stage 1 classifier layer, (iii) adding richer pre-payment signals (e.g., contractual terms, invoice category/type, payment terms, dispute flags, approval workflow signals), (iv) extending and cleaning historical linkages, and (v) explicitly modeling extreme dilution outliers with dedicated tail-handling procedures.

Additionally, ai1 Team has adapted the models to photonic computing environments, such as Microsoft Analogue Optical Computer, Lumai and Ubitech , and expecting comparison results with currently used GPU – based digital environments in the nearest future.

## Acknowledgments

The authors thank the underwriting and risk, data-science and financial engineering teams at The Interface Financial Group for domain expertise and data preparation, and the AI1 Technologies engineering team for model development and infrastructure support.